%% file: 1.main.tex
\documentclass[sigconf]{acmart}
\renewcommand\footnotetextcopyrightpermission[1]{}
\AtBeginDocument{%
  }

\usepackage{booktabs}
\usepackage{multirow}
\usepackage{graphicx}
\usepackage{xcolor} 
\usepackage{array}  
\usepackage{hyperref}
\usepackage{siunitx} 
\usepackage{dsfont} 
\usepackage{subfigure}
\usepackage{multirow}
\usepackage{array}

\usepackage{threeparttable}
\usepackage{graphicx}
\usepackage{textcomp}
\usepackage{xcolor}
\usepackage{enumitem}
\usepackage{comment}
\usepackage{pifont}
\usepackage{amsthm}
\usepackage{algorithm}
\usepackage{amsmath}
\usepackage{bm}
\usepackage{bbding}
\usepackage{amsthm}
\usepackage{algpseudocode}
\usepackage{fontawesome}

\usepackage{amsmath}
\usepackage{amsthm}

\usepackage{dsfont}
\usepackage{booktabs}
\usepackage{multirow}
\usepackage{graphicx}
\usepackage[table]{xcolor}
\usepackage{booktabs}
\usepackage{tabularx}

\definecolor{rank1}{gray}{0.6}  
\definecolor{rank2}{gray}{0.7}  
\definecolor{rank3}{gray}{0.8}  
\definecolor{rank4}{gray}{0.9} 

\newcommand{\first}[1]{\cellcolor{rank1}\textbf{#1}}
\newcommand{\secondc}[1]{\cellcolor{rank2}\underline{#1}}
\newcommand{\third}[1]{\cellcolor{rank3}#1}
\newcommand{\fourth}[1]{\cellcolor{rank4}#1}

\setcopyright{acmlicensed}
\copyrightyear{2018}
\acmYear{2018}
\acmDOI{XXXXXXX.XXXXXXX}
\acmConference[Conference acronym 'XX]{Make sure to enter the correct
  conference title from your rights confirmation email}{June 03--05,
  2018}{Woodstock, NY}
\acmISBN{978-1-4503-XXXX-X/2018/06}

\begin{document}


\title{R${^2}$Energy: A Large-Scale Benchmark for Robust Renewable Energy Forecasting under Diverse and Extreme Conditions}





\author{
  Zhi Sheng\textsuperscript{1}, 
  Yuan Yuan\textsuperscript{2}, 
  Guozhen Zhang\textsuperscript{3},
  Yong Li\textsuperscript{1}
}



\affiliation{%
  \institution{\textsuperscript{1}Center for Urban Science and Computation, Tsinghua University}
  \institution{\textsuperscript{2}Courant Institute of Mathematical Sciences, New York University}
  \institution{\textsuperscript{3}TsingRoc.ai}
  \country{} 
}

\renewcommand{\shortauthors}{Sheng et al.}
\fancyhead[L]{R²Energy: A Large-Scale Benchmark for Robust Renewable
Energy Forecasting}
\begin{abstract}

The rapid expansion of renewable energy, particularly wind and solar power, has made reliable forecasting critical for power system operations. While recent deep learning models have achieved strong average accuracy, the increasing frequency and intensity of climate-driven extreme weather events pose severe threats to grid stability and operational security.
Consequently, developing robust forecasting models that can withstand volatile conditions has become a paramount challenge.
In this paper, we present R$^2$Energy, a large-scale benchmark for NWP-assisted renewable energy forecasting. It comprises over 10.7 million high-fidelity hourly records from 902 wind and solar stations across four provinces in China, providing the diverse meteorological conditions necessary to capture the wide-ranging variability of renewable generation.
We further establish a standardized, leakage-free forecasting paradigm that grants all models identical access to future Numerical Weather Prediction (NWP) signals, enabling fair and reproducible comparison across state-of-the-art representative forecasting architectures.
Beyond aggregate accuracy, we incorporate regime-wise evaluation with expert-aligned extreme weather annotations, uncovering a critical ``robustness gap'' typically obscured by average metrics. This gap reveals a stark robustness-complexity trade-off: under extreme conditions, a model's reliability is driven by its meteorological integration strategy rather than its architectural complexity.
R$^2$Energy provides a principled foundation for evaluating and developing forecasting models for safety-critical power system applications.
\end{abstract}

\received{20 February 2007}
\received[revised]{12 March 2009}
\received[accepted]{5 June 2009}

\maketitle

\input{2.Intro}
\input{3.Related}

\input{4.Data}

\input{5.Experiment}

\input{6.Conclusion}

\bibliographystyle{ACM-Reference-Format}
\bibliography{8.reference}


\end{document}

%% file: 2.Intro.tex
\section{Introduction}
\begin{figure}
    \centering
    \includegraphics[width=1\linewidth]{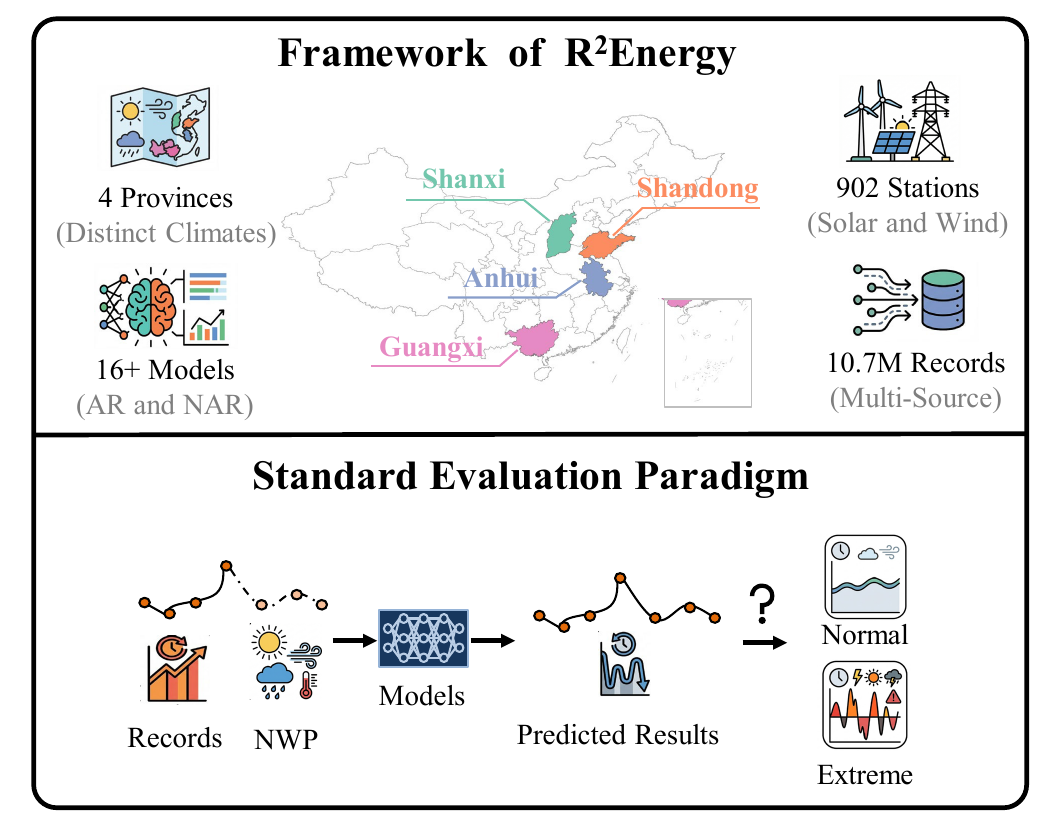}
    \caption{Overall framework of R$^2$Energy.}
    \vspace{-10mm}
    \label{fig:intro}
\end{figure}
The global imperative for carbon neutrality has accelerated the deployment of renewable energy sources, particularly wind and solar photovoltaic (PV) power~\cite{zheng2025strategies,zhu4986631impact,lei2023co,liu2022potential,yuan2022race}. 
In China, the ``Dual Carbon'' goals, aiming for Carbon Peaking by 2030 and Carbon Neutrality by 2060, have driven installed capacity to unprecedented levels~\cite{zhu2024can,wang2025multi}.
However, the inherent intermittency and volatility of renewable energy increasingly jeopardize power system stability as penetration rises~\cite{zheng2025strategies,zhu4986631impact,xu2024resilience,zheng2024climate,zhao2024impacts}. Consequently, modern grid operations, such as unit commitment and dispatch, now rely on accurate, look-ahead forecasting~\cite{xu2025cross,olson2019improving,blaga2019current}, necessitating a paradigm shift from classical time-series extrapolation to Numerical Weather Prediction  (NWP)-assisted, cross-domain modeling~\cite{olson2019improving,ma2024fusionsf,boussif2023improving,yang2023wind}.

Meanwhile, climate change introduces a critical paradox: the energy sources designed to mitigate climate risks are themselves highly sensitive to intensifying weather variability~\cite{xu2024resilience,zhu4986631impact,zheng2024climate,liu2023climate}. As extreme events such as torrential rainstorms and prolonged heatwaves become the "new normal," power systems face heightened operational stress~\cite{zheng2024climate,liu2023climate,zhu4986631impact}. In these regimes, grid stability depends not on average-case accuracy, but on a model's ability to anticipate abrupt ramp-downs and efficiency losses. Developing NWP-informed forecasting frameworks that remain robust under these extreme conditions is therefore essential for reliable resource allocation and the security of safety-critical power systems~\cite{zhu4986631impact,zheng2024climate}.

Despite the rapid proliferation of deep learning architectures~\cite{shrestha2019review,li2024deep,wang2019review}, ranging from RNNs~\cite{sherstinsky2020fundamentals} and Transformers~\cite{vaswani2017attention,boussif2023improving} to advanced MLPs~\cite{shao2022spatial,zeng2023transformers} and generative models~\cite{durgadevi2021generative,yang2024survey}, the academic community continues to evaluate renewable energy forecasting primarily through aggregated metrics like Mean Absolute Error (MAE)~\cite{ma2024fusionsf,boussif2023improving,xu2025cross,ma2022hybrid}. This average-case evaluation paradigm obscures a critical vulnerability: model fragility under extreme meteorological conditions. 
A forecasting system that achieves a marginal reduction in overall error yet fails catastrophically during an abrupt storm front or extreme heat event poses substantial risks to power grid security and operational reliability~\cite{shen2020critical,shen2025quantifying,perera2020quantifying,hawker2024management}.
Two fundamental obstacles hinder progress: 
(i) the reliance on isolated, limited datasets that lack diverse climatic regimes, 
and (ii) the absence of standardized protocols for incorporating future NWP signals, which leads to inherently unfair comparisons between autoregressive and non-Autoregressive models.
Consequently, current benchmarks fail to distinguish models that are ``good on average'' from those that remain reliable under the high-impact, low-frequency events critical for safety-critical power grid operations~\cite{xu2025cross,alessandrini2017characterization,sperati2015weather,pandvzic2023advances,pombo2022benchmarking}.
There is an urgent need for a benchmark that moves beyond aggregate statistics to rigorously evaluate forecasting reliability across heterogeneous climatic regions and meteorological extremities.

To bridge the robustness-evaluation gap identified above, we present \textbf{R$^2$Energy} (benchmark for \underline{\textbf{R}}obust \underline{\textbf{R}}newable \underline{\textbf{Energy}} forecasting). This large-scale, multi-source benchmark evaluates forecasting reliability across two orthogonal dimensions: climatic diversity across distinct geographical regions and meteorological extremity during critical weather events.
Moving beyond simple time-series extrapolation, R$^2$Energy redefines renewable forecasting as an NWP-assisted mapping process. Formally, the task learns the mapping $(X, C) \rightarrow \hat{y}$ within a standardized, leakage-free protocol that ensures all models have identical access to future exogenous signals, enabling fair and reproducible ``stress testing''.
Crucially, to address model reliability under stress, we move beyond aggregate error metrics by introducing the Qualification Rate ($Q$) to measure industrial safety compliance.

As shown in Figure~\ref{fig:intro}, built upon 10.7 million high-fidelity hourly records from 902 wind and solar stations across four diverse climatic regions in China (Anhui, Shanxi, Shandong, and Guangxi), R$^2$Energy provides the unprecedented scale required to evaluate models against high-impact, tail-risk events.
By benchmarking 16 state-of-the-art models through a regime-wise lens, we uncover a critical robustness-complexity trade-off typically obscured by previous studies~\cite{zeng2023transformers,xu2025cross}: performance is driven less by architectural complexity than by a model’s meteorological integration strategy.
Empirically, Autoregressive models, specifically GRU, prove remarkably stable, as their dynamic injection of future NWP signals effectively anchors recursive trajectories and mitigates classical error accumulation.
Conversely, Transformer variants exhibit heightened sensitivity and instability in noisy regimes, such as volatile wind fields. These findings highlight why protocol-aligned robustness is non-negotiable for the deployment of safety-critical power systems.
Our contributions are summarized as follows:
\begin{itemize}[leftmargin=*]
    \item We present R$^2$Energy, a massive benchmark comprising 10.7M records from 902 stations. We leverage this scale to evaluate 16 state-of-the-art models under diverse operational conditions.
    \item  We establish a standardized, leakage-free forecasting protocol that redefines renewable forecasting as an NWP-assisted mapping task. Beyond standard error metrics, we introduce the Qualification Rate ($\mathcal{Q}$) for industrial safety compliance.
    \item We propose a regime-wise, extreme-aware evaluation paradigm using expert-aligned weather annotations that rigorously assess model reliability under non-linear distribution shifts and high-impact tail risks.
    \item Our analysis uncovers a critical robustness-conplexity trade-off, revealing that a model's reliability under pressure is primarily driven by its meteorological integration strategy rather than its architectural complexity.
\end{itemize}

%% file: 3.Related.tex
\section{Preliminaries}

In this section, we define the notation and the problem formulation for the NWP-assisted renewable energy forecasting task. We denote scalars in normal font (e.g., $y$), vectors in bold lowercase (e.g., $\mathbf{x}$), and tensors/sets in calligraphic or capital bold font (e.g., $\mathcal{D}, \mathbf{X}$).

\noindent \paragraph{\textbf{Data Representation.}} We consider a dataset $\mathcal{D} = \{ \mathbf{S}^{(i)} \}_{i=1}^{N_{\text{station}}}$ collected from $N_{\text{station}}$ renewable energy stations. At time $t$, the system state of station $i$ is described by the capacity factor
\begin{equation}
    y_t^{(i)} = \frac{P_{out,t}^{(i)}}{P_{rated}^{(i)}} \in [0,1],
\end{equation}
together with historical covariates $\mathbf{x}_t^{(i)} \in \mathbb{R}^{d_x}$ and a future context vector $\mathbf{c}_t^{(i)} \in \mathbb{R}^{d_c}$ derived from NWP forecasts and deterministic temporal features, which are known \textit{a priori} for the forecasting horizon. Data quality is indicated by a binary mask $M_t^{(i)} \in \{0,1\}$, where $M_t^{(i)}=1$ denotes missing or anomalous observations.

\noindent \paragraph{\textbf{Sample Construction.}}
Using a sliding window with history length $L$ and forecasting horizon $H$, we construct samples $\xi_k^{(i)} = (\mathbf{X}_k^{(i)}, \mathbf{C}_k^{(i)}, \mathbf{Y}_k^{(i)})$, where
\begin{align}
    \mathbf{X}_k^{(i)} &= \big[(y_t^{(i)}, \mathbf{x}_t^{(i)})\big]_{t=t_k-L+1}^{t_k}
    \in \mathbb{R}^{L \times (1+d_x)}, \\
    \mathbf{C}_k^{(i)} &= \big[\mathbf{c}_t^{(i)}\big]_{t=t_k+1}^{t_k+H}
    \in \mathbb{R}^{H \times d_c}, \\
    \mathbf{Y}_k^{(i)} &= \big[y_t^{(i)}\big]_{t=t_k+1}^{t_k+H}
    \in \mathbb{R}^{H \times 1}.
\end{align}
To ensure the benchmark evaluates forecasting performance rather than imputation, a sample is retained only if all involved time steps are valid:
\begin{equation}
    \sum_{t=t_k-L+1}^{t_k+H} M_t^{(i)} = 0.
\end{equation}

\noindent \paragraph{\textbf{NWP-Assisted power Forecasting.}}
Given a historical observation tensor $\mathbf{X} \in \mathbb{R}^{L \times (1+d_x)}$ and the future exogenous context $\mathbf{C} \in \mathbb{R}^{H \times d_c}$ derived from NWP forecasts, the goal is to predict the future capacity factor trajectory $\mathbf{Y} \in \mathbb{R}^{H \times 1}$ over a horizon $H$. We employ a parametric model $f_\theta$ to approximate the mapping $\hat{\mathbf{Y}} = f_\theta(\mathbf{X}, \mathbf{C})$. The optimal parameter set $\theta^*$ is determined by minimizing the expected discrepancy between the predicted and ground-truth trajectories over the valid sample space $\Omega_{\text{valid}}$:
\begin{equation}
    \theta^* = \arg\min_{\theta} \mathbb{E}_{(\mathbf{X}, \mathbf{C}, \mathbf{Y}) \in \Omega_{\text{valid}}}
    \big[ \ell(f_\theta(\mathbf{X}, \mathbf{C}), \mathbf{Y}) \big].
\end{equation}

%% file: 4.Data.tex
\section{The \textbf{R}$^2$Energy Framework}
Reliable evaluation of renewable energy forecasting models hinges on both the scale of available data and the diversity of underlying physical regimes. In this section, we introduce \textbf{$\textbf{R}^2$Energy}, a large-scale, multi-source benchmark designed to capture the heterogeneity of real-world renewable generation across space, time, and extreme operating conditions.

\label{sec:framework}

\input{Tables/Dataset}

\begin{figure*}[t]
    \centering
    \includegraphics[width=0.8\linewidth]{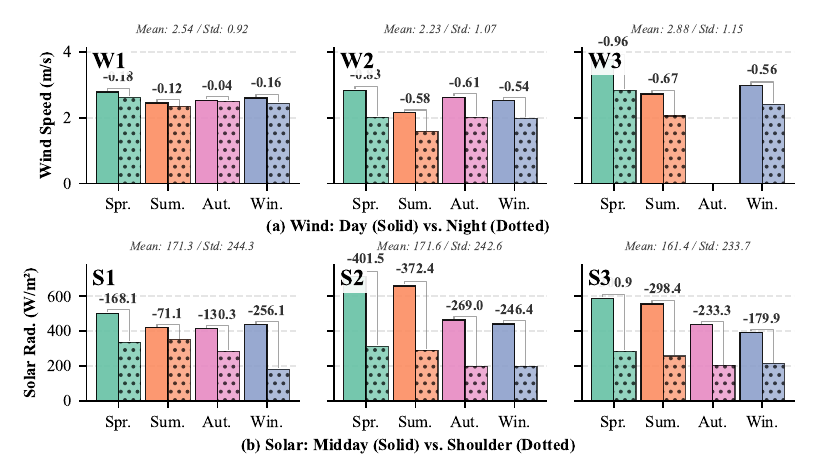}
    \caption{Intra-day and seasonal variability across the wind and solar datasets. The top row (a) illustrates the average wind speed (m/s) variations between daytime (07:00–18:00, solid bars) and nighttime (dotted bars). The bottom row (b) compares solar radiation (W/m²) during peak midday hours (11:00–14:00, solid bars) versus shoulder hours (morning/evening, dotted bars). The numerical labels above the brackets denote the difference between the two periods, while the global mean and standard deviation are listed for each dataset. Missing bars in W3 indicate data unavailability for specific seasons.}
    \label{fig:data}
\end{figure*}

\subsection{Dataset Overview}
~\label{sec:data}
In this section, we introduce the R$^2$Energy dataset in detail and analyze the characteristics of different datasets as well as the impacts of extreme weather events.

\paragraph{\textbf{Renewable Energy Station Data}} We have constructed a massive multi-source dataset comprising over 10.7 million hourly records, covering both wind and solar power generation across four distinct climatic regions in China: Anhui, Shanxi, Shandong, and Guangxi. As detailed in Table~\ref{tbl:dataset_stats_uniform}, this dataset includes 902 individual stations, offering a high degree of geographic diversity and scale compared to previous benchmarks. To standardize the prediction target across stations with varying capacities—and to account for dynamic capacity changes due to potential expansion—we utilize the \textit{Capacity Factor} (CF) as the ground truth. The CF is calculated by dividing the raw power output by the real-time rated capacity of the station. 


\paragraph{\textbf{Meteorological Data}}
Reliable meteorological information is critical for mapping weather conditions to power output. We utilize the ERA5-Land reanalysis dataset~\cite{store2024era5}, which offers a high spatial resolution of $0.1^\circ$ and an hourly temporal resolution. We acquire site-specific meteorological conditions for each station via bilinear interpolation. In this benchmark, these ERA5 variables serve a dual purpose: they act as historical covariates and, crucially, as a high-fidelity proxy for future \textit{Numerical Weather Prediction} (NWP) signals. While operational NWP forecasts contain inherent errors, using reanalysis data as a "perfect prognosis" proxy isolates the forecasting model's mapping capability from the noise of weather prediction errors. This allows us to rigorously evaluate whether a model can effectively learn the physical relationship between weather and power generation.

\paragraph{\textbf{Distribution Characteristics.}} To explicitly characterize the distribution diversity of the target variables, we analyzed the spatiotemporal patterns of Wind Speed and Solar Radiation, the primary meteorological drivers of power generation. As illustrated in Figure \ref{fig:data}, the datasets exhibit distinct distributional shifts across different climatic zones. The wind datasets cover a spectrum from stable to highly volatile regimes. As shown in Figure \ref{fig:data}(a), W1 (Anhui) exhibits a "steady-state" characteristic with moderate wind speeds ($\sim 2.5$ m/s) and minimal diurnal amplitude (day-night differences range only from $-0.04$ to $-0.18$ m/s). In stark contrast, W2 (Shanxi) and W3 (Shandong) represent "high-volatility" regimes typical of northern plains and coastal areas. Specifically, W3 demonstrates extreme diurnal fluctuations, with a sharp drop of $-0.96$ m/s during spring nights, while W2 maintains consistently high variability across all seasons (e.g., $-0.61$ m/s in autumn). This diversity challenges forecasting models to adapt to both persistent patterns (W1) and rapid, high-amplitude transitions (W2/W3). Similarly, the solar datasets reveal significant differences in both intensity and seasonal phase. S2 (Shanxi) typifies a high-altitude, arid climate with intense radiation (Spring peak $>600$ W/m²) and massive intra-day variability, evidenced by the steepest midday-to-shoulder drop of $-401.5$ W/m². Conversely, S3 (Guangxi), located in a humid subtropical zone, shows a distinct seasonal phase shift (peaking in Summer rather than Spring) and significantly lower radiation variance during Spring ($+0.9$ W/m² difference), likely attributed to the persistent cloud cover and rainfall characteristic of the southern rainy season. S1 (Anhui) serves as a transitional distribution with moderate intensity. R$^2$Energy represents more than a mere expansion of data volume; it encompasses a comprehensive range of physical modalities. By covering diverse climatic regimes, our benchmark enables a more holistic evaluation of model generalization potential across varying physical constraints.

\begin{figure}[h]
    \centering
    \includegraphics[width=\linewidth]{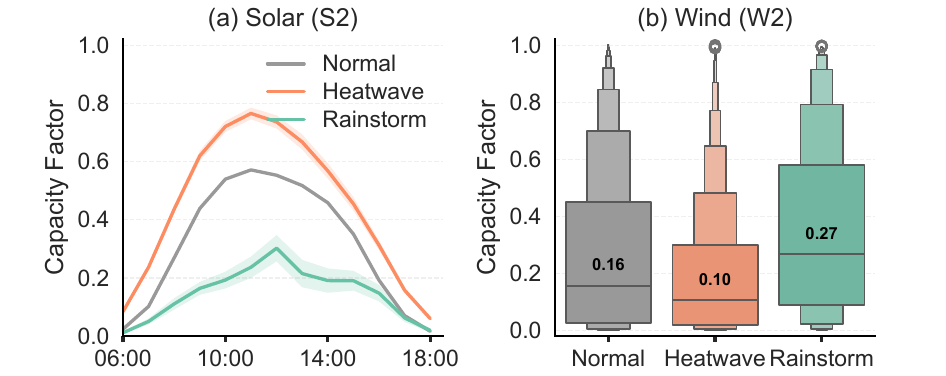}
    \caption{Impact of extreme weather on renewable power generation. (a) Solar (S2) and (b) Wind (W2) datasets. Shaded areas in (a) denote 95\% confidence intervals; values in (b) indicate medians.}
    \label{fig:extreme_impact}
\end{figure}

\paragraph{\textbf{Extreme Weather Signals}}
To diagnose model reliability under stress, we annotate the timeline with extreme weather labels based on the official standards defined by the China Meteorological Administration (CMA). We focus on two high-impact categories: \textit{Rainstorm} and \textit{Heatwave}, each graded into three severity levels. Detailed information are shown in Table~\ref{tab:extreme_definitions}. To empirically validate the necessity of these annotations, we analyzed the power response characteristics under different meteorological states using the Shanxi (S2/W2) datasets. Figure \ref{fig:extreme_impact} visualizes the aggregated diurnal curves for solar power and distributional box-plots for wind power during labeled Level 1-3 events versus normal periods. The results reveal profound physical divergences driven by extreme weather. In the solar domain (Figure \ref{fig:extreme_impact}a), rainstorms induce a drastic suppression effect, reducing the peak capacity factor to below 0.3—less than 50\% of the normal baseline—due to thick cloud occlusion. Conversely, heatwaves coincide with high-irradiance clear-sky conditions, pushing the peak output above 0.75, significantly higher than the normal mean. In the wind domain (Figure \ref{fig:extreme_impact}b), we observe a clear regime-dependent distributional shift. Rainstorms, typically driven by convective systems, introduce stronger wind forcing and elevated turbulence, which shifts the operating distribution toward higher generation (median: 0.16 → 0.27), while heatwaves, often associated with stagnant high-pressure systems, lead to "wind droughts" with a reduced median of 0.10. These significant non-linear shifts confirm that extreme weather imposes distinct physical modalities on power generation. Consequently, forecasting models optimized solely for "average" conditions may fail catastrophically during these high-impact events. R$^2$Energy addresses this critical gap by providing the precise annotations required to evaluate model robustness specifically under these tail-risk regimes.

\input{Tables/extrem}

\subsection{Data Quality Control}
\label{subsec:qc}

Real-world operational data inevitably contains sensor failures and transmission anomalies. To ensure the benchmark evaluates physical modeling capability rather than data cleaning skills, we implement a strict quality-masking strategy. We generate a binary mask $M \in \{0,1\}$ for the power series based on three rigorous criteria: \textbf{(i) Missing Values:} Any timestamp containing NaN values is immediately flagged. \textbf{(ii) Frozen Value Detection:} We identify periods where the power output remains mathematically identical for more than 24 hours, which typically indicates sensor sticking or data logger failures. \textbf{(iii) Statistical Outlier Removal:} We calculate the Z-score of the power series. Data points deviating from the station's mean by more than a threshold $\theta=3.0$ standard deviations are flagged as statistical anomalies.During the sample construction process, we adopt a ``zero-tolerance'' policy: any sample window containing even a single masked time step ($M=1$) is strictly discarded. This approach differs from prior works that rely on interpolation, ensuring that our models are trained and evaluated solely on high-confidence, physically consistent observations.


\subsection{Benchmarking Protocols}
To ensure a fair and consistent comparison between diverse models, we standardize the injection mechanism for future NWP data, adapting the methodologies described in previous work~\cite{xu2025cross}. 

\paragraph{\textbf{Autoregressive (AR) Models}}
For AR models (e.g., LSTM~\cite{hochreiter1997long}, GRU~\cite{chung2014empirical}), which generate predictions iteratively, the NWP data is inserted dynamically at each future time step. Specifically, during the forecasting phase, the NWP vector $\mathbf{c}_{t}$ corresponding to the target timestamp is concatenated with the model's input (or hidden state) for that specific iteration. This simulates the operational availability of weather forecasts for each specific look-ahead hour.

\paragraph{\textbf{Non-Autoregressive (NAR) Models}}
For NAR models, which predict the entire horizon simultaneously, the injection method depends on the specific architecture: \textbf{(i) General NAR Models (e.g., CNN/MLP-based):} The future NWP sequence is typically treated as a known covariate channel. It is concatenated with the encoded hidden state vector or the input features, serving as a global condition for the decoder to generate the full sequence. \textbf{(ii) Transformer-based Models (e.g., Informer~\cite{zhou2021informer}):} The NWP sequence is integrated into the generative inference process. The NWP data is zero-padded to match the dimensions of the decoder's token input and then concatenated. This combined input is then processed through the embedding layers (Value, Position, and Temporal Embeddings) before being fed into the decoder, ensuring the attention mechanism can explicitly attend to future weather contexts.

%% file: Tables/Dataset.tex
\begin{table*}[t!]
\caption{Statistics of the six datasets. The rightmost column lists the variables and settings shared by all datasets.}
\label{tbl:dataset_stats_uniform}
\resizebox{\textwidth}{!}{
\begin{tabular}{lllc c r rr l}
\toprule
\multirow{2}{*}{\textbf{Type}} & \multirow{2}{*}{\textbf{Dataset}} & \multirow{2}{*}{\textbf{Province}} & \multirow{2}{*}{\textbf{\# of Stations}} & \multirow{2}{*}{\textbf{Time Period}} & \multirow{2}{*}{\textbf{Total Records}} & \multicolumn{2}{c}{\textbf{Extreme (Lvl 1-3)}} & \multirow{2}{*}{\textbf{Variables \& Settings}} \\
\cmidrule(lr){7-8}
 &  &  &  &  &  & \textbf{Rainstorm} & \textbf{Heatwave} &  \\
\midrule

\multirow{3}{*}{\textbf{Wind}} 
 & \textbf{W1} & Anhui & 110 & 2023-01-01 -- 2024-11-18 & 1,548,258 & 6,723 & 56,839 & \multirow{6}{*}{\parbox{4.5cm}{
    \textbf{Physical Vars:}\\ 
    $ws_{10}, ssrd, u_{10}, v_{10}, t_{2m}, tp, CF$ \\[3pt]
    \textbf{Quality Control:}\\ 
    Masks for Missing \& Outliers \\[3pt]
    \textbf{Extreme Labels:}\\ 
    Rainstorm \& Heatwave (Lvl 1-3)
 }} \\
 & \textbf{W2} & Shanxi & 233 & 2023-10-30 -- 2025-03-25 & 2,324,213 & 1,882 & 3,953 &  \\
 & \textbf{W3} & Shandong & 20 & 2024-12-01 -- 2025-07-14 & 82,752 & 0 & 2,164 &  \\
 
\cmidrule{1-8} 

\multirow{3}{*}{\textbf{Solar}} 
 & \textbf{S1} & Anhui & 242 & 2023-01-01 -- 2024-11-18 & 3,545,797 & 13,771 & 135,802 &  \\
 & \textbf{S2} & Shanxi & 241 & 2023-10-30 -- 2024-12-31 & 2,362,101 & 2,434 & 7,848 &  \\
 & \textbf{S3} & Guangxi & 56 & 2023-01-01 -- 2025-01-01 & 902,917 & 899 & 978 &  \\

\bottomrule
\end{tabular}
}
\end{table*}

%% file: Tables/extrem.tex

\begin{table}[t]
\centering
\caption{Definition of extreme meteorological events and severity levels used in RexBench.}
\label{tab:extreme_definitions}
\resizebox{\columnwidth}{!}{
\begin{tabular}{l c c l}
\toprule
\textbf{Event Type} & \textbf{Level} & \textbf{Warning} & \textbf{Criteria} \\
\midrule
\multirow{3}{*}{Rainstorm} 
 & Level 1 & Blue   & Rainfall $\geq 50$ mm within 12 h \\
 & Level 2 & Yellow & Rainfall $\geq 50$ mm within 6 h  \\
 & Level 3 & Orange & Rainfall $\geq 50$ mm within 3 h  \\
\midrule
\multirow{3}{*}{Heatwave} 
 & Level 1 & Yellow & Max temperature $> 35^\circ$C for 3 consecutive days \\
 & Level 2 & Orange & Max temperature $> 37^\circ$C within 24 h \\
 & Level 3 & Red    & Max temperature $> 40^\circ$C within 24 h \\
\bottomrule
\end{tabular}
}
\end{table}

%% file: 5.Experiment.tex
\section{Experimental Settings}
In this section, we detail the experimental setup of R$^2$Energy, including the task formulations, baseline models, evaluation metrics, and implementation protocols.

\input{Tables/task}

\subsection{Operational Forecasting Regimes}
\label{sec:regimes}
In power system operations, renewable energy forecasting is categorized by standardized temporal horizons, each associated with distinct modeling assumptions and operational objectives~\cite{teixeira2024advancing,aslam2021survey,natarajan2019survey}. Table~\ref{tab:forecasting_horizons} summarizes the commonly adopted forecasting regimes in the energy industry. R$^2$Energy focuses on forecasting horizons within the USTF, STF, and MTF regimes, where future Numerical Weather Prediction (NWP) information is routinely available and operationally utilized. In all settings, historical observations $X_{t-L+1:t}$ and future NWP availability $C_{t+1:t+H}$ are utilized to predict the target $Y_{t+1:t+H}$:\textbf{(i) Ultra-Short-Term Forecasting (USTF):} Defined as using a historical window of $L=24$ hours to predict the next $H=1$ hour. \textbf{(ii) Short-Term Forecasting (STF):} Defined as using $L=24$ hours of history to predict the next $H=24$ hours (Day-Ahead). \textbf{(iii) Medium-Term Forecasting (MTF):} Defined as using $L=72$ hours of history to predict the next $H=72$ hours. This is critical for maintenance scheduling and congestion management.

\subsection{Baseline Selection}
To rigorously benchmark the proposed dataset, we evaluate a comprehensive suite of 16 representative models across three distinct categories: Statistical/Physical methods, Autoregressive (AR) models, and Non-Autoregressive (NAR) models.

\begin{figure}[h]
    \centering
    \includegraphics[width=0.85\linewidth]{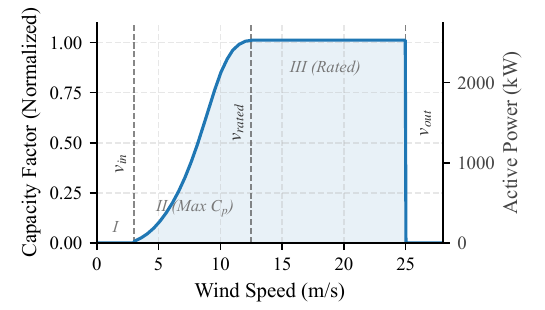} 
    \caption{Power curve of the GE 2.5MW wind turbine. The curve illustrates the non-linear relationship between wind speed and power generation, highlighting three operational regions: cut-in ($v_{in} \approx 3$ m/s), rated ($v_{rated} \approx 12.5$ m/s), and cut-out ($v_{out} = 25$ m/s).}
    \label{fig:power_curve}
\end{figure}

\paragraph{\textbf{Statistical and Physical Baselines.}} 
These methods serve as foundational benchmarks, representing physics-agnostic heuristics and domain-knowledge-driven estimations.
\textbf{Seasonal-Naive:} Given the diurnal periodicity of renewable energy, we use the observation from the same timestamp of the previous day as the prediction: $\hat{y}_{t+h} = y_{t+h-24}$ (recursively applied for horizons $H>24$).
\textbf{Physics-Informed Models:} We implement classic physical conversion models~\cite{zheng2025strategies} that map meteorological variables (subsets of the context vector $\mathbf{c}_t$) to capacity factors.
For \textbf{Solar Power}, let $R_t$ be the solar irradiance ($W/m^2$) and $T_t$ be the ambient air temperature ($^\circ C$) derived from $\mathbf{c}_t$. The estimated capacity factor $\hat{y}_t$ is calculated as:
\begin{equation}
    \hat{y}_t =  \frac{R_t}{R_{std}} \times \left[ 1 + \alpha_p (T_t - T_{std}) \right],
\end{equation}
where standard test conditions are set to $R_{std}=1000 W/m^2$ and $T_{std}=25^\circ C$. The temperature coefficient is set to $\alpha_p = -0.0035/^\circ C$. 
For \textbf{Wind Power}, we utilize the power curve of the widely deployed GE 2.5MW wind turbine. The wind speed at hub height ($v_{hub}$) is first extrapolated from the 10m wind speed ($v_{10m} \in \mathbf{c}_t$) using the power law with a shear exponent $\alpha=0.143$: $v_{hub} = v_{10m} \cdot (h_{hub}/10)^\alpha$, where $h_{hub}=100m$. The power output is then interpolated from the manufacturer's power curve (Figure~\ref{fig:power_curve}).

\paragraph{\textbf{Autoregressive (AR) Models.}}
These models generate predictions iteratively ($\hat{y}_{t+1} = f(y_t, h_t, \mathbf{c}_{t+1})$) and are suitable for capturing sequential dependencies. We evaluate \textbf{RNN Variants} including standard \textbf{RNN}~\cite{elman1990finding}, \textbf{LSTM}~\cite{hochreiter1997long}, and \textbf{GRU}~\cite{chung2014empirical}, along with their bidirectional counterparts~\cite{schuster1997bidirectional,hochreiter1997long,chung2014empirical} (\textbf{Bi-LSTM}, \textbf{Bi-GRU}). Future NWP data is injected at each decoding step.

\paragraph{\textbf{Non-Autoregressive (NAR) Models.}}
These models predict the entire horizon simultaneously and are widely considered State-of-the-Art (SOTA) in long-term time series forecasting. We evaluate three sub-categories: \textbf{(i) CNN-based:} \textbf{TCN}~\cite{bai2018empirical} and \textbf{TimesNet}~\cite{wu2022timesnet}, leveraging 1D or 2D convolutions to capture temporal variations; \textbf{(ii) MLP-based:} \textbf{DLinear}~\cite{zeng2023transformers}, \textbf{STID}~\cite{shao2022spatial}, and \textbf{TSMixer}~\cite{chen2023tsmixer}, which prioritize efficiency; and \textbf{(iii) Transformer-based:} \textbf{Autoformer}~\cite{wu2021autoformer}, \textbf{Informer}~\cite{zhou2021informer}, and \textbf{iTransformer}~\cite{liu2023itransformer}, employing attention mechanisms to model global weather contexts.

\subsection{Evaluation Metrics}
To provide a comprehensive assessment, we employ standard statistical metrics, including Mean Absolute Error (MAE) and Root Mean Square Error (RMSE). However, purely statistical errors often fail to reflect the operational viability of forecasts in real-world power grids. Therefore, we adopt the Qualification Rate ($\text{Q}$), a standard industrial metric used to evaluate whether prediction errors remain within the safety tolerance thresholds required for grid dispatch and stability. Furthermore, absolute error metrics are inherently sensitive to the magnitude of the target variable. This magnitude dependence makes it difficult to fairly compare model robustness across different meteorological regimes. To address this limitation and explicitly quantify the value of our model against Physics-based baselines, we propose the Forecast Skill Score ($\text{S}$).

\paragraph{\textbf{Industrial Qualification Rate ($\text{Q}$).}}
Following power system standards, we utilize the Qualification Rate to measure the proportion of predictions that satisfy operational safety margins. Let $\tau$ denote the tolerance threshold (set to $\tau=0.25$ in our experiments). The rate $\text{Q}$ is defined as:
\begin{equation}
   \text{Q} = \frac{1}{|\Omega_{\text{test}}|} \sum_{(t, i) \in \Omega_{\text{test}}} \mathbb{I}\left( |y_t^{(i)} - \hat{y}_t^{(i)}| \le (1 - \tau) \right) \times 100\%,
\end{equation}
where $\mathbb{I}(\cdot)$ is the indicator function, and $\Omega_{\text{test}}$ represents the set of all valid test samples. A higher $\text{Q}$ indicates better compliance with grid dispatch requirements.

\paragraph{\textbf{Forecast Skill Score ($\text{S}$).}}
To explicit quantify the value of our model against established baselines across different scenarios, we propose the Forecast Skill Score. This metric measures the percentage improvement of the deeplearning models over a benchmark baseline (e.g., Physics-based or Persistence methods), defined as:
\begin{equation}
    \text{S} = \left( 1 - \frac{\text{MAE}_{\text{model}}}{\text{MAE}_{\text{baseline}}} \right) \times 100\%.
\end{equation}
A positive $\text{S}$ indicates the model outperforms the baseline, while a negative value implies degradation. Comparing $S_{\text{norm}}$ and $S_{\text{ext}}$ reveals whether the model maintains its predictive advantage under stress. Importantly, we select the \emph{best-performing physics-based baseline} to ensure a conservative and industry-relevant evaluation. Specifically, we use physical conversion models for solar power forecasting and the standard GE 2.5MW turbine power curve for wind power forecasting.

\subsection{Experimental Settings}
All experiments are implemented in PyTorch and conducted on NVIDIA GPUs. To ensure reproducibility and prevent information leakage, each dataset is strictly split by time into Training, Validation, and Test sets with a ratio of 7:1:2. The target variable, power generation capacity factor, is naturally bounded within $[0,1]$, while all meteorological covariates are standardized using Z-score normalization computed exclusively on the training set. Models are trained for a maximum of 20 epochs using the AdamW optimizer with a weight decay of $1\times10^{-5}$, and early stopping is applied with a patience of three epochs based on validation loss. We consider commonly used ranges for key hyperparameters, including learning rate $\in \{10^{-1}, \dots, 10^{-5}\}$ and dropout $\in \{0, 0.05, 0.1\}$ for all models; autoregressive models (RNN/LSTM/GRU) additionally vary the number of layers in $[1,2]$ and hidden size in $[128,256]$; CNN and MLP models vary the number of layers in $[2,4]$ and model dimension $d_{\text{model}} \in [32,64,128]$; and Transformer-based models vary encoder/decoder layers in $[1,2]$ with $d_{\text{model}} \in [32,64]$. For each model, we select the best-performing configuration. Each experiment is repeated five times with different random seeds, and we report the mean and standard deviation of all evaluation metrics.

\input{Tables/S-Res}

\input{Tables/W-Res}

\section{Experimental Results and Analysis}
\label{sec:main_results}

\subsection{Overall Performance}
In this section, we present a comprehensive analysis of the forecasting performance across 16 models on the R$^2$Energy datasets, focusing on the overall hierarchy, temporal sensitivity, modality characteristics, and operational reliability. The results are shown in Table~\ref{tbl:solar-main} and Table~\ref{tbl:wind-main}.


\paragraph{\textbf{Overall Performance Hierarchy.}}

A macro-level inspection of in Table~\ref{tbl:solar-main} and Table~\ref{tbl:wind-main} reveals a clear hierarchy: deep learning models substantially outperform statistical and physics-based baselines. For example, in the Solar S1 dataset (Panel A), the best-performing deep learning models achieve an MAE of approximately 0.052, representing a reduction of over 50\% compared to the physical baseline (MAE 0.1283). However, beyond this expected gap, R$^2$Energy exposes a fundamental trend: performance is governed less by model complexity and more by the alignment between the architecture and the data-generating process. Across both wind and solar datasets, Autoregressive (AR) models, particularly GRU and Bi-GRU, emerge as consistent leaders. In the ultra-short-term forecast (Horizon=1), GRU achieves the lowest MAE on the W1 (0.0567), S1 (0.0527), and S2 (0.0541) datasets. This advantage stems from their decoding paradigm. In R$^2$Energy, renewable generation is treated as an NWP-assisted mapping task where meteorological covariates are available at every step. AR models inject these signals dynamically during decoding, preserving a tight temporal coupling between instantaneous weather forcing and power output. In contrast, Transformer-based models, which rely on global attention, often struggle to filter high-frequency stochastic fluctuations in low signal-to-noise regimes. Consequently, models like Autoformer significantly underperform simpler RNN variants in wind datasets (e.g., W1 MAE 0.2078 vs. GRU 0.0567).

\paragraph{\textbf{Impact of Forecasting Horizons.}}
The relative strengths of model families evolve systematically with forecasting horizon, revealing how NWP availability reshapes classical horizon-dependent trade-offs. (i) \textbf{Ultra-Short-Term (USTF, 1h):} In the immediate reaction regime, AR models dominate across nearly all datasets (e.g., GRU ranking first in W1, S1–S3). At this horizon, the most recent system state $y_t$ remains highly informative, and recursive decoding benefits from strong short-term autocorrelation. The superiority of AR models here reflects their ability to directly exploit local temporal continuity; (ii) \textbf{Short-Term (STF, 24h):} Contrary to the conventional expectation that AR models suffer from cumulative error, GRU-based models maintain their leading positions in the day-ahead setting. This observation is critical: future NWP signals act as an external correction mechanism, effectively anchoring the recursive trajectory to physically plausible weather evolutions. As a result, error propagation is substantially mitigated, and AR models retain stability even over 24 steps; (iii) \textbf{Medium-Term (MTF, 72h):} At longer horizons, performance gaps narrow. While AR models remain competitive, Non-Autoregressive (NAR) architectures such as TimesNet and TCN become increasingly viable, particularly in volatile wind datasets (e.g., surpassing LSTM in W3). This transition suggests that when immediate historical states lose predictive power, models that emphasize global temporal patterns and multi-periodicity gain relative advantage. R$^2$Energy thus highlights a regime-dependent crossover rather than a single universally optimal architecture.

The relative strengths of model families evolve systematically with the forecasting horizon.
(i) \textbf{Ultra-Short-Term (1h):} In the immediate reaction regime, AR models dominate. For instance, in Table~\ref{tbl:solar-main}, GRU and RNN variants consistently rank first across S1–S3. At this horizon, the most recent system state $y_t$ is highly informative, and recursive decoding exploits local temporal continuity effectively.
(ii) \textbf{Short-Term (24h):} Contrary to the expectation that AR models suffer from error accumulation, GRU-based models maintain their lead. In Table~\ref{tbl:wind-main} (Panel B), GRU achieves an MAE of 0.1062 on W1, significantly outperforming the best Transformer model (iTransformer, MAE 0.1212). This stability suggests that future NWP signals act as an external correction mechanism, anchoring the recursive trajectory to physically plausible weather evolutions.
(iii) \textbf{Medium-Term (72h):} At longer horizons, performance gaps narrow. While AR models remain competitive, Non-Autoregressive (NAR) architectures like TimesNet begin to show advantages in volatile datasets (e.g., W3), where they capture multi-scale temporal patterns that recursive methods may miss.


\paragraph{\textbf{Modality-Specific Characteristics.}}
R$^2$Energy validates that "Wind" and "Solar" constitute distinct physical modalities requiring different modeling capabilities. Solar Power (S1–S3) operates in a pattern-dominant regime. Governed by deterministic diurnal cycles and direct irradiance relationships, solar datasets exhibit high predictability, reflected in consistently high qualification rates (Q > 90\% for top models). In this setting, most competent deep learning models converge to similar performance ceilings, and architectural sophistication yields diminishing returns. Temporal consistency and accurate phase alignment, rather than expressive capacity, become the dominant factors. Wind Power (W1–W3), by contrast, represents a noise-dominant regime. As characterized in Section~\ref{sec:data}, datasets such as W2 and W3 exhibit strong diurnal asymmetry and high variance driven by turbulence and synoptic transitions. Here, overall Q-scores drop to 60–80\%, and model rankings compress significantly. The relative improvement of CNN-based models (e.g., TCN) in wind tasks suggests that local receptive fields are effective for capturing short-lived gust structures and regime shifts, a property less critical in solar forecasting. This modality gap underscores why benchmarking across both wind and solar is essential for assessing generalizable forecasting robustness.


\paragraph{\textbf{Training Stability and Industrial Reliability.}}
Beyond average accuracy, R$^2$Energy emphasizes training stability as a first-class evaluation dimension for safety-critical deployment. The reported standard deviations across five independent runs reveal stark contrasts between model families. Lightweight architectures, particularly GRU, Bi-GRU, and MLP-based models, exhibit highly stable convergence surfaces, with Std values typically below 0.002 in solar datasets. Such determinism is crucial for operational environments where retraining and redeployment are routine. In contrast, Transformer-based models demonstrate pronounced volatility, especially in high-noise wind regimes (e.g., Autoformer exhibiting Std = 0.254 in W3 USTF). This instability indicates a rugged optimization landscape exacerbated by stochastic meteorological inputs and limited effective signal. From an industrial perspective, such variance poses tangible risks: inconsistent model behavior under retraining undermines trust and complicates system integration. Collectively, these results suggest that robustness in renewable energy forecasting is not synonymous with architectural complexity, and that stability-aware benchmarks like R$^2$Energy are necessary to bridge the gap between academic performance and operational reliability.


\subsection{Robustness Under Extreme Weather}

\input{Tables/extrem_res}
In this section, we conduct a fine-grained diagnostic analysis of model robustness under extreme meteorological regimes. We focus on the Short-Term Forecasting task ($H=24$) for Solar (S1) and Wind (W1) datasets(shown in Table~\ref{tbl:extreme-main}). To differentiate between "high-impact" and "high-difficulty" scenarios, we categorize events by their entropic nature and report three metrics: \textbf{MAE} (physical error), \textbf{Qualification Rate ($Q$)} (compliance with industrial safety margins), and \textbf{Forecast Skill Score ($S$)} (improvement relative to the physics-based baseline).

\paragraph{\textbf{Solar Power (S1): The Low-Entropy Extreme.}}
The results on S1 reveal a "Clear-Sky Paradox." During \textbf{Heatwave} events, we observe a universal performance improvement compared to the Normal regime. For instance, the leading GRU model reduces its MAE from 0.0871 (Normal) to 0.0680 (Heatwave), with the Qualification Rate ($Q$) approaching near-perfect levels (97.51\%). This improvement highlights that heatwaves, while operationally critical, represent a \textit{low-entropy} forecasting task: they are typically accompanied by stable, cloud-free conditions. In this deterministic regime, Deep Learning models successfully capture the non-linear temperature-derating coefficients, significantly outperforming the physical Baseline (Skill Score $\mathbf{S}=45.68\%$). Conversely, \textbf{Rainstorm} events introduce a \textit{high-entropy} regime dominated by dynamic cloud occlusion. While absolute MAE values decrease due to suppressed total output, the relative Skill Scores expose architectural fragilities. The Baseline fails to track rapid cloud transients. Under these conditions, Autoregressive (AR) models (RNN/GRU) maintain robustness ($\mathbf{S} \approx 10-31\%$, $Q > 97\%$). In stark contrast, Transformer-based architectures exhibit instability; the Autoformer degrades to a negative Skill Score ($\mathbf{S} = -26.06\%$), performing worse than the primitive physical baseline. This suggests that global attention mechanisms, while effective for semantic context, struggle to model the local, high-frequency "shading" effects inherent to rainstorms.

\paragraph{\textbf{Wind Power (W1): Turbulence and the Robustness Gap.}}
The Wind (W1) dataset presents a rigorous stress test characterized by non-stationary regime shifts. The distinction between model classes becomes stark under Rainstorm conditions (Panel B), where the physical Baseline suffers a catastrophic collapse (MAE: 0.2720, $Q$: 27-49\%). This underscores the inability of static power curves to account for convective turbulence. In this high-noise regime, we identify a decisive "Robustness Gap." The GRU and Bi-GRU models demonstrate exceptional resilience (MAE $\approx 0.15$, $Q \approx 80\%$), achieving a massive Skill Score improvement of over 40\%. Notably, the TCN also performs strongly ($\mathbf{S}=44.96\%$), suggesting that architectures with local receptive fields are well-suited for capturing short-lived gust features. However, the Autoformer collapses ($S = -78.57\%$, $Q = 31.26\%$). This indicates that during high-entropy wind events, global correlation modeling tends to interpret high-frequency turbulence as noise to be smoothed, leading to under-dispersed and operationally dangerous predictions.

\paragraph{\textbf{Diagnostic Summary: Alignment of Inductive Bias.}}
The cross-comparison leads to a critical finding: architectural complexity does not correlate with operational robustness in NWP-driven tasks. Simple, lightweight recurrent architectures (GRU/RNN) consistently outperform complex Transformers in high-stakes weather scenarios. We attribute this to the alignment between the model's inductive bias and the physical task: power generation during a storm is a causal, step-wise reaction to immediate meteorological forcing. AR models, which inject NWP data dynamically at each decoding step, mirror this physical causality. Conversely, generative Non-Autoregressive models appear prone to \textit{signal dilution}, where the precise NWP forcing at time $t$ is washed out by the global attention mechanism, rendering them less reliable for safety-critical dispatch.

\begin{figure}
    \centering
    \includegraphics[width=0.8\linewidth]{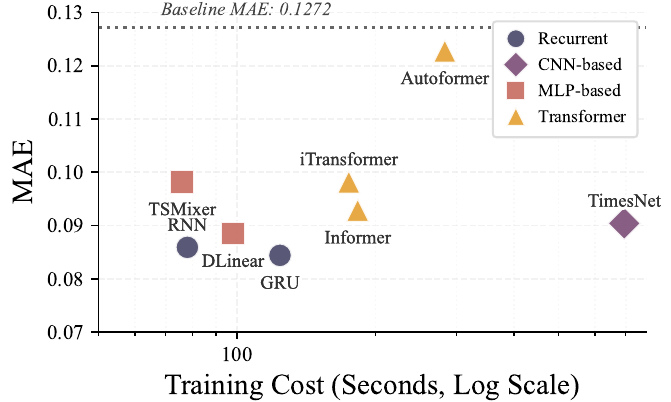}
    \caption{Efficiency-Accuracy Landscape (Solar S1, H=24). The scatter plot illustrates the trade-off between computational cost (Training Time per Epoch, Log Scale) and predictive performance (MAE). The dashed line represents the physical Baseline error (0.1272).}
\label{fig:efficiency}
\end{figure}

\subsection{Operational Efficiency}
Beyond predictive accuracy, the deployment of forecasting models in real-time grid dispatch systems is constrained by strict latency budgets and computational resource limits. Figure \ref{fig:efficiency} presents a cost–benefit analysis on the Solar (S1) dataset ($H=24$), plotting MAE against per-epoch training cost (seconds, log scale). The results challenge the prevailing trend toward increasingly deep architectures. The bottom-left quadrant of the efficiency landscape, representing the optimal trade-off between speed and accuracy, is dominated by lightweight Recurrent (GRU, RNN) and MLP-based (DLinear, TSMixer) models. Notably, the GRU achieves the lowest error ($\text{MAE} \approx 0.084$) while requiring nearly an order of magnitude less training cost than CNN-based alternatives. DLinear and TSMixer further demonstrate strong efficiency, training in under 100 seconds with competitive accuracy, making them well suited for resource-constrained deployment. In contrast, Transformer models (Informer, iTransformer, Autoformer) cluster in the upper-middle region, exhibiting a clear negative computational return. Despite incurring $2\times$–$5\times$ higher training costs than the GRU, they deliver inferior accuracy. The Autoformer is a notable outlier: its costly auto-correlation mechanisms fail to yield meaningful gains, achieving an MAE ($\approx 0.122$) only marginally better than the naive Baseline. This indicates that the quadratic or log-linear complexity of attention mechanisms is poorly matched to renewable energy time series, where dynamics are primarily governed by local physical causality rather than long-range semantic dependencies. Overall, these findings suggest that, for operational renewable energy forecasting, \textit{less is often more}, and that structurally aligned AR and MLP models offer a more compute-efficient path than increasingly over-parameterized architectures.


%% file: Tables/task.tex
\begin{table}[t]
\centering
\caption{Taxonomy of Renewable Energy Forecasting Horizons and Applications.}
\label{tab:forecasting_horizons}
\resizebox{\columnwidth}{!}{
\begin{tabular}{l c p{4cm}}
\toprule
\textbf{Forecast Type} & \textbf{Horizon} & \textbf{Operational Use Cases} \\
\midrule
Ultra-Short-Term (USTF) & $\leq$ 1 hour 
& Real-time dispatch,\\
& & Automatic Generation Control,\\
& & Power Smoothing. \\
\midrule
Short-Term (STF) & $\leq$ 1 day 
& Economic Dispatch,\\
& & Reserve Scheduling,\\
& & Day-ahead Market Operations. \\
\midrule
Medium-Term (MTF) & 1--15 days 
& Unit Commitment,\\
& & Maintenance Planning,\\
& & Congestion Management. \\
\midrule
Long-Term (LTF) & Months to years 
& Capacity Planning,\\
& & Grid Expansion,\\
& & Risk Assessment. \\
\bottomrule
\end{tabular}
}
\end{table}

    
        
        
        
        
        

%% file: Tables/S-Res.tex
\begin{table*}[t!]
\centering
\caption{Prediction results on three Datasets (S1, S2, S3). The Baseline corresponds to a physics-informed photovoltaic formulation and the Seasonal-Naive method, respectively. The tasks are categorized into three Panels based on forecast horizons. Results are formatted as \textbf{Mean (Std)}. Darker background indicates better performance (Rank 1 to 4).}
\label{tbl:solar-main}

\resizebox{\textwidth}{!}{
\begin{tabular}{l!{\color{black}\vrule} ccc !{\color{black}\vrule} ccc !{\color{black}\vrule} ccc}
\toprule
\multirow{2}{*}{\textbf{Model}}
& \multicolumn{3}{c!{\color{black}\vrule}}{\textbf{S1}}
& \multicolumn{3}{c!{\color{black}\vrule}}{\textbf{S2}}
& \multicolumn{3}{c}{\textbf{S3}} \\
\cmidrule(lr){2-4} \cmidrule(lr){5-7} \cmidrule(lr){8-10}
 & \textbf{MAE} $\downarrow$ & \textbf{RMSE} $\downarrow$ & \textbf{Q (\%)} $\uparrow$
 & \textbf{MAE} $\downarrow$ & \textbf{RMSE} $\downarrow$ & \textbf{Q (\%)} $\uparrow$
 & \textbf{MAE} $\downarrow$ & \textbf{RMSE} $\downarrow$ & \textbf{Q (\%)} $\uparrow$ \\
\midrule

\multicolumn{10}{c}{\textit{\textbf{Panel A: Ultra-short-term Forecast (Horizon = 1)}}} \\
\midrule
\textbf{Baseline} & 0.1283/0.1318 & 0.1725/0.2151 & 86.41/80.89 & 0.1760/0.1545 & 0.2346/0.2490 & 72.33/76.39 & 0.1166/0.1274 & 0.1717/0.2054 & 86.75/82.40 \\
RNN          & \secondc{0.0527} (0.001)  & \third{0.0866} (0.000) & \third{97.70} (0.026)           & \third{0.0576} (0.003)  & \third{0.0964} (0.003)  & \third{96.78} (0.217)           & \third{0.0570} (0.001)  & \third{0.0888} (0.001)  & \third{97.57} (0.047) \\
LSTM         & 0.2170 (0.135)          & 0.2997 (0.174)          & 69.83 (22.70)           & 0.2034 (0.124)          & 0.2930 (0.164)          & 71.52 (20.78)           & 0.1744 (0.098)          & 0.2536 (0.137)          & 76.18 (17.56) \\
Bi-LSTM      & 0.1628 (0.134)          & 0.2289 (0.174)          & 79.12 (22.72)           & 0.1541 (0.123)          & 0.2268 (0.164)          & 80.01 (20.78)           & 0.1348 (0.098)          & 0.1981 (0.136)          & 83.35 (17.56) \\
GRU          & \first{0.0527} (0.001)  & \first{0.0864} (0.000)  & \first{97.72} (0.010)   & \secondc{0.0541} (0.002) & \secondc{0.0930} (0.002) & \secondc{96.96} (0.091)  & \secondc{0.0564} (0.001) & \secondc{0.0880} (0.001) & \secondc{97.61} (0.049) \\
Bi-GRU       & \third{0.0528} (0.001) & \secondc{0.0866} (0.000) & \secondc{97.71} (0.017)  & \first{0.0533} (0.000)  & \first{0.0917} (0.001)  & \first{97.08} (0.037)   & \first{0.0554} (0.001)  & \first{0.0871} (0.000)  & \first{97.67} (0.022) \\
TCN          & 0.2270 (0.000)          & 0.3416 (0.000)          & 61.30 (0.000)           & 0.2052 (0.121)          & 0.2941 (0.163)          & 71.50 (20.75)           & 0.1753 (0.097)          & 0.2544 (0.136)          & 76.18 (17.56) \\
TimesNet     & 0.0610 (0.002)          & 0.0949 (0.002)          & 97.30 (0.107)           & \fourth{0.0644} (0.002) & 0.1046 (0.002)          & 96.22 (0.210)           & \fourth{0.0599} (0.001) & \fourth{0.0916} (0.001) & 97.44 (0.069) \\
DLinear      & 0.1670 (0.101)          & 0.2212 (0.126)          & 76.31 (21.23)           & 0.1602 (0.057)          & 0.2074 (0.061)          & 78.31 (12.58)           & 0.1747 (0.044)          & 0.2370 (0.054)          & 74.32 (10.58) \\
STID         & 0.1096 (0.109)          & 0.1603 (0.141)          & 88.20 (18.45)           & 0.0685 (0.005)          & 0.1075 (0.006)          & 96.00 (0.516)           & 0.0963 (0.079)          & 0.1445 (0.110)          & 90.35 (14.26) \\
TSMixer      & {0.0595} (0.002) & {0.0927} (0.002) & \fourth{97.42} (0.171)  & 0.0650 (0.001)          & \fourth{0.1042} (0.002) & \fourth{96.30} (0.319)  & 0.0607 (0.002)          & 0.0920 (0.002)          & \fourth{97.45} (0.164) \\
Autoformer   & 0.1335 (0.006)          & 0.1837 (0.007)          & 81.78 (1.579)           & 0.1706 (0.109)          & 0.2428 (0.151)          & 78.67 (19.70)           & 0.1565 (0.080)          & 0.2260 (0.114)          & 80.17 (14.97) \\
Informer     & 0.0838 (0.003)          & 0.1236 (0.002)          & 94.01 (0.178)           & 0.1311 (0.087)          & 0.1921 (0.118)          & 85.07 (15.26)           & 0.0736 (0.002)          & 0.1090 (0.003)          & 95.86 (0.308) \\
iTransformer & \fourth{0.0580} (0.001)  & \fourth{0.0926} (0.001)  & 97.29 (0.080)           & 0.0666 (0.001)          & 0.1070 (0.002)          & 95.53 (0.178)           & 0.0616 (0.000)          & 0.0931 (0.001)          & 97.26 (0.068) \\
\midrule

\multicolumn{10}{c}{\textit{\textbf{Panel B: Short-term Forecast (Horizon = 24)}}} \\
\midrule
\textbf{Baseline} & 0.1272/0.1303 & 0.1702/0.2127 & 86.79/81.12 & 0.1731/0.1530 & 0.2305/0.2464 & 72.91/76.61 & 0.1143/0.1247 & 0.1683/0.2010 & 87.25/82.86 \\
RNN          & \third{0.0859} (0.001)  & \secondc{0.1303} (0.000) & \secondc{93.63} (0.036)  & \secondc{0.1032} (0.001) & \third{0.1560} (0.001) & \fourth{89.01} (0.253)  & \third{0.0928} (0.001)  & \third{0.1373} (0.001)  & \third{91.78} (0.239) \\
LSTM         & 0.1359 (0.096)          & 0.1942 (0.124)          & 84.82 (16.89)           & \fourth{0.1035} (0.002)  & \fourth{0.1577} (0.001) & \third{89.04} (0.316)   & 0.1227 (0.064)          & 0.1801 (0.090)          & 86.13 (11.90) \\
Bi-LSTM      & 0.1844 (0.118)          & 0.2565 (0.152)          & 76.35 (20.66)           & 0.1842 (0.099)          & 0.2654 (0.132)          & 75.10 (16.85)           & 0.1542 (0.079)          & 0.2251 (0.111)          & 80.15 (14.55) \\
GRU          & \first{0.0844} (0.001)  & \first{0.1297} (0.001)  & \first{93.69} (0.032)   & \first{0.1022} (0.001)  & \first{0.1550} (0.000)  & \first{89.41} (0.124)   & \first{0.0889} (0.001)  & \first{0.1333} (0.000)  & \first{92.36} (0.039) \\
Bi-GRU       & \secondc{0.0858} (0.001) & \fourth{0.1319} (0.001) & \fourth{93.32} (0.198)  & \third{0.1035} (0.002)  & \secondc{0.1560} (0.001) & \secondc{89.30} (0.236)  & \secondc{0.0893} (0.001) & \secondc{0.1339} (0.001) & \secondc{92.23} (0.086) \\
TCN          & 0.0910 (0.001)          & 0.1354 (0.001)          & 92.84 (0.083)           & 0.1146 (0.002)          & 0.1675 (0.001)          & 86.75 (0.310)           & 0.0987 (0.001)          & 0.1430 (0.001)          & 91.13 (0.176) \\
TimesNet     & 0.0904 (0.001)          & 0.1359 (0.001)          & 92.67 (0.169)           & 0.1180 (0.005)          & 0.1733 (0.008)          & 85.87 (0.788)           & 0.0975 (0.000)          & 0.1417 (0.001)          & 91.28 (0.098) \\
DLinear      & \fourth{0.0885} (0.003) & \third{0.1306} (0.003)  & \third{93.40} (0.477)   & 0.1259 (0.011)          & 0.1767 (0.012)          & 84.10 (2.731)           & \fourth{0.0943} (0.003) & \fourth{0.1388} (0.004) & \fourth{91.69} (0.507) \\
STID         & 0.1878 (0.115)          & 0.2607 (0.149)          & 75.60 (20.11)           & 0.1730 (0.069)          & 0.2454 (0.096)          & 75.84 (11.53)           & 0.1290 (0.061)          & 0.1859 (0.087)          & 85.55 (11.62) \\
TSMixer      & 0.0982 (0.007)          & 0.1438 (0.009)          & 91.18 (1.631)           & 0.1321 (0.011)          & 0.1904 (0.017)          & 82.37 (2.613)           & 0.1023 (0.004)          & 0.1485 (0.006)          & 90.11 (1.013) \\
Autoformer   & 0.1228 (0.007)          & 0.1729 (0.007)          & 85.34 (1.792)           & 0.1657 (0.070)          & 0.2316 (0.098)          & 76.67 (11.17)           & 0.1045 (0.003)          & 0.1518 (0.001)          & 90.09 (0.093) \\
Informer     & 0.0929 (0.001)          & 0.1361 (0.001)          & 92.11 (0.125)           & 0.1734 (0.066)          & 0.2379 (0.095)          & 76.20 (10.91)           & 0.1023 (0.002)          & 0.1500 (0.002)          & 90.08 (0.272) \\
iTransformer & 0.0982 (0.001)          & 0.1425 (0.001)          & 90.87 (0.157)           & 0.1238 (0.002)          & 0.1746 (0.001)          & 84.66 (0.291)           & 0.0946 (0.000)          & 0.1389 (0.000)          & 91.55 (0.031) \\
\midrule

\multicolumn{10}{c}{\textit{\textbf{Panel C: Medium-term Forecast (Horizon = 72)}}} \\
\midrule
\textbf{Baseline} & 0.1273/0.3262 & 0.1694/0.4254 & 86.91/49.05 & 0.1698/0.3126 & 0.2260/0.4208 & 73.63/51.82 & 0.1128/0.2489 & 0.1661/0.3467 & 87.53/61.86 \\
RNN          & \third{0.0861} (0.003)  & \fourth{0.1311} (0.002) & \fourth{93.59} (0.326)  & \fourth{0.1077} (0.005)  & \fourth{0.1612} (0.004)  & \fourth{88.05} (1.061)   & \third{0.0918} (0.001) & \third{0.1358} (0.002) & \third{91.70} (0.356) \\
LSTM         & 0.0875 (0.001)          & 0.1332 (0.001)          & 93.33 (0.154)           & \third{0.1047} (0.001) & \third{0.1610} (0.002) & \third{88.68} (0.283)  & 0.1217 (0.062)          & 0.1786 (0.088)          & 86.14 (11.38) \\
Bi-LSTM      & 0.1880 (0.123)          & 0.2609 (0.155)          & 75.52 (21.32)           & 0.1880 (0.123)          & 0.2609 (0.155)          & 75.52 (21.32)           & 0.1524 (0.076)          & 0.2231 (0.107)          & 80.29 (13.81) \\
GRU          & \secondc{0.0858} (0.003) & \third{0.1309} (0.002)  & \third{93.69} (0.278)   & \secondc{0.1003} (0.001)  & \secondc{0.1563} (0.002)  & \secondc{89.62} (0.317)   & \secondc{0.0906} (0.002)  & \first{0.1337} (0.001)  & \first{92.04} (0.135) \\
Bi-GRU       & {0.0893} (0.001) & {0.1383} (0.001)  & {92.23} (0.198)   & \first{0.0990} (0.001)  & \first{0.1545} (0.000)  & \first{89.88} (0.145)   & \first{0.0898} (0.001)  & \secondc{0.1342} (0.001)  & \secondc{91.88} (0.283) \\
TCN          & 0.0917 (0.001)          & 0.1362 (0.001)          & 92.69 (0.161)           & 0.1230 (0.001)          & 0.1773 (0.001)          & 84.86 (0.335)           & 0.1022 (0.001)          & 0.1444 (0.000)          & 90.75 (0.083) \\
TimesNet     & 0.0920 (0.001)          & 0.1367 (0.001)          & 92.69 (0.185)           & {0.1216} (0.001) & 0.1772 (0.001)          & \fourth{85.10} (0.111)  & 0.0996 (0.001)          & 0.1450 (0.001)          & 90.45 (0.223) \\
DLinear      & \fourth{0.0874} (0.003) & \secondc{0.1291} (0.003) & \secondc{93.84} (0.450)  & 0.1270 (0.007)          & 0.1756 (0.008)          & 84.16 (2.190)           & 0.0990 (0.003)          & 0.1414 (0.005)          & \fourth{91.16} (0.672) \\
STID         & 0.2897 (0.097)          & 0.3893 (0.123)          & 57.90 (17.02)           & 0.1759 (0.065)          & 0.2485 (0.090)          & 74.35 (10.03)           & 0.1605 (0.069)          & 0.2296 (0.102)          & 79.60 (13.26) \\
TSMixer      & 0.0975 (0.010)          & 0.1436 (0.012)          & 91.25 (2.436)           & 0.1338 (0.010)          & 0.1919 (0.015)          & 81.57 (2.387)           & 0.1124 (0.012)          & 0.1600 (0.016)          & 87.67 (3.457) \\
Autoformer   & 0.1126 (0.003)          & 0.1580 (0.003)          & 87.83 (0.673)           & 0.1689 (0.069)          & 0.2328 (0.098)          & 75.31 (10.53)           & 0.1003 (0.002)          & 0.1478 (0.001)          & 90.13 (0.124) \\
Informer     & \first{0.0838} (0.003)  & \first{0.1236} (0.002)  & \first{94.01} (0.178)   & 0.1501 (0.078)          & 0.2170 (0.106)          & 81.32 (13.50)           & \fourth{0.0957} (0.001)  & \fourth{0.1397} (0.001)  & {91.08} (0.267) \\
iTransformer & 0.0987 (0.001)          & 0.1420 (0.001)          & 91.13 (0.214)           & 0.1295 (0.002)          & 0.1799 (0.002)          & 82.61 (0.486)           & {0.0958} (0.001) & {0.1405} (0.001) & 91.00 (0.163) \\

\bottomrule
\end{tabular}
}
\end{table*}

%% file: Tables/W-Res.tex
\begin{table*}[t!]
\centering
\caption{Prediction results on three Datasets (W1, W2, W3). The Baseline corresponds to a turbine power-curve-based formulation and the Seasonal-Naive method, respectively. The tasks are categorized into three Panels based on forecast horizons. Results are formatted as \textbf{Mean (Std)}. Darker background indicates better performance (Rank 1 to 4).}
\label{tbl:wind-main}

\resizebox{\textwidth}{!}{
\begin{tabular}{l!{\color{black}\vrule} ccc !{\color{black}\vrule} ccc !{\color{black}\vrule} ccc}
\toprule
\multirow{2}{*}{\textbf{Model}}
& \multicolumn{3}{c!{\color{black}\vrule}}{\textbf{W1}}
& \multicolumn{3}{c!{\color{black}\vrule}}{\textbf{W2}}
& \multicolumn{3}{c}{\textbf{W3}} \\
\cmidrule(lr){2-4} \cmidrule(lr){5-7} \cmidrule(lr){8-10}
 & \textbf{MAE} $\downarrow$ & \textbf{RMSE} $\downarrow$ & \textbf{Q (\%)} $\uparrow$
 & \textbf{MAE} $\downarrow$ & \textbf{RMSE} $\downarrow$ & \textbf{Q (\%)} $\uparrow$
 & \textbf{MAE} $\downarrow$ & \textbf{RMSE} $\downarrow$ & \textbf{Q (\%)} $\uparrow$ \\
\midrule

\multicolumn{10}{c}{\textit{\textbf{Panel A: Ultra-short-term Forecast (Horizon = 1)}}} \\
\midrule
\textbf{Baseline} & 0.2042/0.2106 & 0.2975/0.2934 & 66.86/- & 0.2544/0.2546 & 0.3650/0.3451 & 61.07/59.58 & 0.1443/0.1747 & 0.2220/0.2480 & 80.01/74.68 \\
RNN          & \fourth{0.0586} (0.001) & \fourth{0.0889} (0.000) & \fourth{97.85} (0.050)           & \fourth{0.0627} (0.001) & \fourth{0.0957} (0.000) & \fourth{97.29} (0.032)  & \first{0.0600} (0.001)  & \first{0.0939} (0.001)  & \secondc{97.31} (0.055) \\
LSTM         & 0.1761 (0.098)          & 0.2536 (0.137)          & 74.64 (13.70)           & 0.2019 (0.114)          & 0.2889 (0.159)          & 72.88 (20.00)           & 0.1398 (0.053)          & 0.2048 (0.080)          & 81.98 (11.93) \\
Bi-LSTM      & 0.1364 (0.097)          & 0.1979 (0.137)          & 80.21 (13.68)           & 0.1555 (0.114)          & 0.2241 (0.159)          & 81.06 (20.01)           & 0.1168 (0.054)          & 0.1693 (0.082)          & 87.15 (12.17) \\
GRU          & \secondc{0.0567} (0.001)  & \first{0.0863} (0.001)  & \first{98.05} (0.060)           & \secondc{0.0618} (0.001) & \secondc{0.0942} (0.001) & \third{97.40} (0.069)   & \secondc{0.0657} (0.001) & \third{0.0984} (0.001)  & \third{97.10} (0.043) \\
Bi-GRU       & \first{0.0567} (0.000)  & \secondc{0.0867} (0.001) & \secondc{98.02}(0.081)           & \first{0.0614} (0.001)  & \first{0.0936} (0.001)  & \first{97.45} (0.054)   & \third{0.0662} (0.001)  & \secondc{0.0969} (0.001) & \first{97.32} (0.053) \\
TCN          & 0.0591 (0.000)          & 0.0893 (0.000)          & \third{97.92} (0.030)   & 0.0638 (0.000)          & 0.0972 (0.000)          & 97.23 (0.034)           & 0.1407 (0.023)          & 0.1829 (0.023)          & 82.53 (5.961) \\
TimesNet     & \third{0.0578} (0.001)  & \third{0.0884} (0.002)  & {97.89} (0.114)  & \third{0.0623} (0.000)  & \third{0.0944} (0.000)  & \secondc{97.43} (0.036)  & \fourth{0.0922} (0.008) & \fourth{0.1336} (0.005) & \fourth{93.38} (0.791) \\
DLinear      & 0.1360 (0.026)          & 0.1822 (0.030)          & 85.27 (4.517)           & 0.0892 (0.051)          & 0.1253 (0.058)          & 92.35 (9.793)           & 0.2018 (0.017)          & 0.2673 (0.013)          & 66.31 (7.949) \\
STID         & 0.1296 (0.066)          & 0.1863 (0.093)          & 84.45 (13.12)           & 0.0645 (0.002)          & 0.0990 (0.003)          & 97.04 (0.185)           & 0.1519 (0.020)          & 0.2181 (0.029)          & 78.84 (4.221) \\
TSMixer      & 0.0662 (0.004)          & 0.1006 (0.007)          & {96.97} (0.640)   & 0.0650 (0.001)          & 0.0986 (0.001)          & 97.15 (0.091)           & 0.1066 (0.017)          & 0.1599 (0.022)          & 88.97 (3.717) \\
Autoformer   & 0.2078 (0.043)          & 0.2993 (0.060)          & 67.50 (7.457)           & 0.1525 (0.001)          & 0.2117 (0.001)          & 78.47 (0.224)           & 0.3092 (0.254)          & 0.3837 (0.229)          & 58.33 (28.06) \\
Informer     & 0.0942 (0.001)          & 0.1363 (0.001)          & 92.17 (0.262)           & 0.0941 (0.001)          & 0.1370 (0.001)          & 92.30 (0.187)           & 0.1428 (0.035)          & 0.1898 (0.039)          & 82.52 (8.518) \\
iTransformer & 0.0888 (0.009)          & 0.1315 (0.014)          & {92.81} (1.842)  & 0.1090 (0.014)          & 0.1529 (0.019)          & 90.00 (2.692)           & 0.1455 (0.019)          & 0.1837 (0.019)          & 82.55 (5.026) \\
\midrule

\multicolumn{10}{c}{\textit{\textbf{Panel B: Short-term Forecast (Horizon = 24)}}} \\
\midrule
\textbf{Baseline} & 0.2023/0.2084 & 0.2950/0.2907 & 67.23/67.38 & 0.2540/0.2538 & 0.3649/0.3441 & 61.20/59.73 & 0.1424/0.1721 & 0.2184/0.2449 & 80.36/75.15 \\
RNN          & \secondc{0.1067} (0.001) & \first{0.1505} (0.001) & \first{89.89} (0.135)           & \third{0.1501} (0.002)  & \fourth{0.2010} (0.002)  & \third{81.37} (0.434)   & \first{0.1221} (0.002)  & \secondc{0.1702} (0.001) & \first{86.66} (0.245) \\
LSTM         & 0.1363 (0.059)          & 0.1937 (0.084)          & 83.78 (11.81)           & \fourth{0.1491} (0.001)          & \secondc{0.1988} (0.000)          & \secondc{81.66} (0.287)           & 0.1516 (0.015)          & 0.2018 (0.032)          & 81.00 (4.260) \\
Bi-LSTM      & 0.1657 (0.072)          & 0.2359 (0.103)          & 77.90 (14.48)           & 0.2073 (0.072)          & 0.2869 (0.108)          & 71.60 (12.18)           & 0.1588 (0.017)          & 0.2181 (0.039)          & 78.77 (5.022) \\
GRU          & \first{0.1062} (0.000)  & \secondc{0.1507} (0.000) & \secondc{89.81} (0.110)           & \first{0.1458} (0.002)  & \first{0.1958} (0.001)  & \first{82.10} (0.333)  & \secondc{0.1251} (0.001) & \first{0.1699} (0.001)  & \secondc{86.43} (0.150) \\
Bi-GRU       & \third{0.1080} (0.000)  & \third{0.1529} (0.001)  & \third{89.48} (0.164)           & \secondc{0.1497} (0.001) & \third{0.1998} (0.001) & \fourth{81.27} (0.262)           & \third{0.1300} (0.003)  & \third{0.1744} (0.002)  & \third{85.33} (0.521) \\
TCN          & {0.1131} (0.000) & {0.1608} (0.000) & \third{88.12} (0.084)   & 0.1580 (0.001)          & 0.2092 (0.001)          & 79.56 (0.142)           & 0.1420 (0.006) & \fourth{0.1857} (0.005) & 82.69 (1.835) \\
TimesNet     & \fourth{0.1115} (0.001)          & \fourth{0.1587} (0.001)          & \fourth{88.52} (0.135)   & 0.1549 (0.001)          & 0.2072 (0.001)          & 80.05 (0.270)           & 0.1475 (0.007)          & 0.1940 (0.006)          & 80.72 (1.521) \\
DLinear      & 0.1178 (0.003)          & 0.1630 (0.005)          & {88.12} (0.639)  & 0.1633 (0.012)          & 0.2124 (0.012)          & 78.28 (2.855)           & 0.1779 (0.044)          & 0.2298 (0.057)          & 75.36 (9.449) \\
STID         & 0.1768 (0.054)          & 0.2497 (0.080)          & 74.69 (10.86)           & 0.1720 (0.011)          & 0.2294 (0.020)          & 76.10 (2.855)           & 0.1879 (0.028)          & 0.2569 (0.041)          & 71.31 (6.736) \\
TSMixer      & 0.1307 (0.013)          & 0.1902 (0.021)          & 83.53 (3.163)           & 0.1593 (0.001)          & 0.2128 (0.003)          & 79.01 (0.307)           & 0.1659 (0.017)          & 0.2437 (0.026)          & 76.17 (3.102) \\
Autoformer   & 0.2119 (0.051)          & 0.3051 (0.070)          & 67.47 (8.944)           & 0.1891 (0.001)          & 0.2634 (0.001)          & 70.77 (0.137)           & 0.1802 (0.000)          & 0.2665 (0.000)          & 72.63 (0.016) \\
Informer     & 0.1292 (0.001)          & 0.1805 (0.002)          & 84.39 (0.295)           & 0.1638 (0.001)          & 0.2160 (0.001)          & 78.18 (0.175)           & \fourth{0.1412} (0.009)          & 0.1902 (0.007)          & \fourth{83.54} (1.689) \\
iTransformer & 0.1212 (0.002)          & 0.1755 (0.002)          & {85.92} (0.351)  & {0.1696} (0.005) & {0.2198} (0.005) & 76.64 (1.047)           & 0.1487 (0.026)          & 0.1955 (0.023)          & 80.39 (6.299) \\
\midrule

\multicolumn{10}{c}{\textit{\textbf{Panel C: Medium-term Forecast (Horizon = 72)}}} \\
\midrule
\textbf{Baseline} & 0.2004/0.2626 & 0.2927/0.3475 & 67.61/57.43 & 0.2499/0.2863 & 0.3612/0.3805 & 61.95/54.97 & 0.1424/0.1955 & 0.2180/0.2690 & 80.34/70.29 \\
RNN          & \third{0.1116} (0.002) & \secondc{0.1575} (0.002) & \third{88.81} (0.390)   & \fourth{0.1628} (0.003) & \fourth{0.2140} (0.003) & \fourth{78.62} (0.694)   & \first{0.1275} (0.005)  & \first{0.1737} (0.003)  & \first{85.83} (0.809) \\
LSTM         & 0.1384 (0.054)          & 0.1959 (0.080)          & 80.41 (7.989)           & \third{0.1582} (0.002)          & \third{0.2097} (0.001)          & \third{79.59} (0.344)           & \fourth{0.1411} (0.006)          & \fourth{0.1833} (0.005)          & \fourth{83.67} (1.323) \\
Bi-LSTM      & 0.1651 (0.067)          & 0.2355 (0.098)          & 76.45 (9.814)           & 0.2101 (0.066)          & 0.2910 (0.102)          & 71.17 (10.92)           & 0.1642 (0.017)          & 0.2255 (0.036)          & 77.02 (4.667) \\
GRU          & \first{0.1088} (0.000)  & \first{0.1536} (0.001)  & \first{89.36} (0.154)   & \first{0.1524} (0.002)  & \first{0.2047} (0.001)  & \first{80.62} (0.294)   & \secondc{0.1322} (0.005) & \secondc{0.1760} (0.004) & \secondc{84.93} (1.052) \\
Bi-GRU       & \secondc{0.1098} (0.001) & \secondc{0.1551} (0.001) & \secondc{89.08} (0.129)  & \secondc{0.1536} (0.003)  & \third{0.2050} (0.003)  & \secondc{80.38} (0.481)  & {0.1447} (0.009) & {0.1892} (0.010) & {81.63} (2.589) \\
TCN          & \fourth{0.1171} (0.001)  & {0.1674} (0.001) & 87.19 (0.203)           & 0.1720 (0.003)          & 0.2267 (0.002)          & 76.39 (0.653)           & 0.1425 (0.014)          & 0.1893 (0.012)          & 81.83 (3.450) \\
TimesNet     & {0.1176} (0.002) & \fourth{0.1673} (0.002)  & 87.00 (0.315)           & {0.1723} (0.003) & 0.2284 (0.002)          & 76.12 (0.704)           & \third{0.1331} (0.005)  & \third{0.1825} (0.005)  & \third{84.05} (1.227) \\
DLinear      & 0.1222 (0.007)          & 0.1676 (0.008)          & \fourth{87.71} (0.957)  & 0.1735 (0.012)          & {0.2226} (0.011) & 76.83 (2.228)           & 0.2054 (0.032)          & 0.2503 (0.045)          & 67.11 (7.259) \\
STID         & 0.1224 (0.005)          & 0.1723 (0.003)          & 86.58 (0.537)           & 0.1786 (0.001)          & 0.2315 (0.000)          & 75.66 (0.105)           & 0.1987 (0.051)          & 0.2847 (0.063)          & 70.34 (8.213) \\
TSMixer      & 0.1575 (0.046)          & 0.2234 (0.066)          & 79.07 (8.922)           & 0.1732 (0.011)          & 0.2337 (0.020)          & 75.73 (2.483)           & 0.1830 (0.029)          & 0.2606 (0.039)          & 72.56 (5.708) \\
Autoformer   & 0.1869 (0.049)          & 0.2694 (0.071)          & 72.04 (8.699)           & 0.2078 (0.041)          & 0.2895 (0.063)          & 69.07 (5.636)           & 0.1756 (0.015)          & 0.2578 (0.021)          & 73.61 (3.130) \\
Informer     & 0.1219 (0.001)          & 0.1747 (0.002)          & 85.71 (0.376)           & 0.1634 (0.001)          & {0.2190} (0.001) & {78.25} (0.254)  & 0.1415 (0.006)          & 0.1875 (0.006)          & 83.73 (1.115) \\
iTransformer & 0.1187 (0.001)          & 0.1722 (0.002)          & 86.70 (0.298)           & 0.1731 (0.003)          & 0.2238 (0.002)          & 75.73 (0.552)           & 0.1448 (0.021)          & 0.1920 (0.019)          & 81.42 (4.999) \\

\bottomrule
\end{tabular}
}
\end{table*}

%% file: Tables/extrem_res.tex
\begin{table*}[t!]
\centering
\caption{Robustness Evaluation under Extreme Weather Conditions (Task: Short-Term Forecasting, H=24). We report the MAE, Qualification Rate (Q), and Forecast Skill Score ( \textbf{S}).  \textbf{S} measures the relative improvement over the Baseline; \textbf{higher  \textbf{S} indicates better performance}. The Baseline corresponds to a physics-informed photovoltaic formulation and the Seasonal-Naive method. Results are formatted as \textbf{Mean (Std)}. Darker background indicates better performance (Rank 1 to 4).}
\label{tbl:extreme-main}

\resizebox{\textwidth}{!}{
\begin{tabular}{l ccc ccc ccc}
\toprule
\multirow{2}{*}{\textbf{Model}}
& \multicolumn{3}{c}{\textbf{Normal}}
& \multicolumn{3}{c}{\textbf{Heatwave}}
& \multicolumn{3}{c}{\textbf{Rainstorm}} \\
\cmidrule(lr){2-4} \cmidrule(lr){5-7} \cmidrule(lr){8-10}
 & \textbf{MAE} $\downarrow$ & \textbf{Q (\%)} $\uparrow$ & \textbf{S (\%)} $\uparrow$
 & \textbf{MAE} $\downarrow$ & \textbf{Q (\%)} $\uparrow$ & \textbf{S (\%)} $\uparrow$
 & \textbf{MAE} $\downarrow$ & \textbf{Q (\%)} $\uparrow$ & \textbf{S (\%)} $\uparrow$ \\
\midrule

\multicolumn{10}{c}{\textit{\textbf{Panel A: Short-term Forecast (Horizon = 24) on S1}}} \\
\midrule
\textbf{Baseline} & 0.1269/0.1397& 86.10/79.36 & 0.00 (--) & 0.1309/0.0697 & 90.67/92.39 & 0.00 (--) &0.0687/0.2064& 95.93/70.46 & 0.00 (--) \\
RNN & \secondc{0.0885} (0.001) & \secondc{92.97} (0.045) & \secondc{30.26} (0.788) & \fourth{0.0711} (0.001) & \first{97.54} (0.105) & \fourth{45.68} (0.764) & \first{0.0615} (0.002) & \first{97.29} (0.310) & \first{10.48} (2.911) \\
LSTM & 0.1350 (0.089) & 84.66 (15.81) & -6.38 (70.13) & 0.1437 (0.145) & 85.36 (24.10) & -9.78 (110.7) & 0.0730 (0.009) & 96.13 (1.990) & -6.26 (13.10) \\
Bi-LSTM & 0.1799 (0.108) & 76.72 (19.33) & -41.77 (85.11) & 0.2168 (0.177) & 73.27 (29.49) & -65.62 (135.2) & 0.0760 (0.012) & 95.16 (2.454) & -10.63 (17.46) \\
GRU & \first{0.0871} (0.001) & \first{93.05} (0.039) & \first{31.36} (0.788) & \first{0.0680} (0.002) & \secondc{97.51} (0.065) & \first{48.05} (1.528) & \secondc{0.0680} (0.004) & \secondc{97.09} (0.363) & \secondc{1.02} (5.822) \\
Bi-GRU & \third{0.0887} (0.001) & \fourth{92.64} (0.218) & \third{30.10} (0.788) & \secondc{0.0685} (0.002) & \third{97.37} (0.068) & \secondc{47.67} (1.528) & \fourth{0.0691} (0.002) & \third{97.03} (0.235) & -0.58 (2.911) \\
TCN & 0.0940 (0.000) & 92.12 (0.091) & 25.93 (0.000) & 0.0727 (0.002) & \fourth{97.28} (0.048) & 44.46 (1.528) & 0.0816 (0.001) & 93.69 (0.151) & -18.78 (1.456) \\
TimesNet & 0.0936 (0.001) & 91.94 (0.158) & 26.24 (0.788) & \third{0.0709} (0.004) & 97.07 (0.374) & \third{45.84} (3.056) & 0.0809 (0.001) & 93.60 (0.230) & -17.76 (1.456) \\
DLinear & \fourth{0.0911} (0.003) & \third{92.78} (0.492) & \fourth{28.21} (2.364) & 0.0730 (0.003) & 97.08 (0.409) & 44.23 (2.292) & 0.0745 (0.006) & 96.53 (0.509) & -8.44 (8.733) \\
STID & 0.1840 (0.105) & 75.95 (18.76) & -45.00 (82.74) & 0.2149 (0.179) & 72.82 (29.11) & -64.17 (136.7) & 0.0904 (0.007) & 92.48 (1.220) & -31.59 (10.19) \\
TSMixer & 0.1018 (0.008) & 90.24 (1.867) & 19.78 (6.304) & 0.0762 (0.005) & 96.87 (0.286) & 41.79 (3.820) & 0.0850 (0.004) & 93.07 (0.718) & -23.73 (5.822) \\
Autoformer & 0.1272 (0.008) & 83.88 (2.165) & -0.24 (6.304) & 0.0968 (0.003) & 94.10 (0.590) & 26.05 (2.292) & 0.0866 (0.001) & 92.93 (0.205) & -26.06 (1.456) \\
Informer & 0.0958 (0.001) & 91.28 (0.150) & 24.51 (0.788) & 0.0762 (0.003) & 97.11 (0.266) & 41.79 (2.292) & \third{0.0688} (0.002) & \fourth{96.76} (0.285) & \third{-0.15} (2.911) \\
iTransformer & 0.1013 (0.001) & 89.99 (0.173) & 20.17 (0.788) & 0.0795 (0.002) & 96.23 (0.138) & 39.27 (1.528) & 0.0795 (0.002) & 94.32 (0.363) & -15.72 (2.911) \\
\midrule
\multicolumn{10}{c}{\textit{\textbf{Panel B: Short-term Forecast (Horizon = 24) on W1}}} \\
\midrule 
\textbf{Baseline} & 0.2023/0.2129 & 67.23/66.47 & 0.00 (--) & 0.1638/0.1573 & 74.37/77.65 & 0.00 (--) & 0.2720/0.4203 & 49.68/27.67 & 0.00 (--) \\
RNN & \secondc{0.1076} (0.001) & \first{89.62} (0.162) & \secondc{46.81} (0.494) & \secondc{0.0962} (0.002) & \third{92.74} (0.109) & \third{41.27} (1.221) & \secondc{0.1510} (0.008) & \first{81.25} (2.743) & \secondc{44.49} (2.941) \\
LSTM & 0.1378 (0.060) & 83.43 (12.06) & 31.88 (29.66) & 0.1154 (0.038) & 88.06 (8.796) & 29.55 (23.20) & 0.2636 (0.194) & 63.37 (26.18) & 3.09 (71.32) \\
Bi-LSTM & 0.1681 (0.073) & 77.40 (14.77) & 16.91 (36.08) & 0.1338 (0.046) & 83.93 (10.99) & 18.32 (28.08) & 0.3628 (0.235) & 49.52 (31.43) & -33.38 (86.39) \\
GRU & \first{0.1071} (0.000) & \secondc{89.52} (0.116) & \first{47.06} (0.000) & \first{0.0953} (0.001) & \first{92.94} (0.151) & \first{41.82} (0.611) & 0.1595 (0.008) & 78.73 (1.605) & 41.36 (2.941) \\
Bi-GRU & \third{0.1091} (0.000) & \third{89.16} (0.167) & \third{46.07} (0.000) & 0.0964 (0.001) & \secondc{92.79} (0.465) & 41.15 (0.611) & \third{0.1529} (0.008) & \fourth{79.69} (1.205) & \third{43.79} (2.941) \\
TCN & 0.1140 (0.000) & 87.90 (0.121) & 43.65 (0.000) & 0.0966 (0.000) & 91.73 (0.134) & 41.03 (0.000) & \first{0.1497} (0.006) & \secondc{80.54} (1.427) & \first{44.96} (2.206) \\
TimesNet & \fourth{0.1130} (0.001) & \fourth{88.18} (0.152) & \fourth{44.14} (0.494) & \first{0.0953} (0.001) & 92.04 (0.039) & \first{41.82} (0.611) & \fourth{0.1533} (0.002) & \third{79.81} (0.512) & \fourth{43.64} (0.735) \\
DLinear & 0.1195 (0.003) & 87.73 (0.644) & 40.93 (1.483) & 0.1018 (0.003) & \fourth{92.17} (0.679) & 37.85 (1.832) & 0.1566 (0.010) & 78.40 (2.178) & 42.43 (3.676) \\
STID & 0.1791 (0.054) & 74.17 (10.93) & 11.47 (26.69) & 0.1475 (0.043) & 80.58 (9.611) & 9.95 (26.25) & 0.3429 (0.214) & 51.25 (28.91) & -26.07 (78.68) \\
TSMixer & 0.1322 (0.013) & 83.19 (3.164) & 34.65 (6.426) & 0.1118 (0.013) & 87.60 (3.217) & 31.75 (7.936) & 0.2258 (0.040) & 65.32 (7.793) & 16.99 (14.71) \\
Autoformer & 0.2154 (0.052) & 66.81 (9.184) & -6.48 (25.70) & 0.1654 (0.031) & 75.34 (5.968) & -0.98 (18.93) & 0.4857 (0.199) & 31.26 (24.42) & -78.57 (73.16) \\
Informer & 0.1314 (0.001) & 83.89 (0.270) & 35.05 (0.494) & 0.1062 (0.002) & 89.61 (0.438) & 35.16 (1.221) & 0.1882 (0.018) & 70.80 (5.120) & 30.81 (6.618) \\
iTransformer & 0.1230 (0.002) & 85.61 (0.239) & 39.20 (0.989) & 0.1050 (0.003) & 89.47 (0.599) & 35.90 (1.832) & 0.1585 (0.015) & 78.34 (3.740) & 41.73 (5.515) \\
\bottomrule
\end{tabular}
}
\end{table*}

%% file: 6.Conclusion.tex
\section{Conclusion}

The introduction of R$^2$Energy marks a paradigm shift in renewable energy forecasting, moving the focus from aggregate statistical precision toward operational robustness under diverse and extreme conditions. Our findings reveal a critical robustness-complexity trade-off, demonstrating that in safety-critical power systems, reliability is driven by the alignment between a model’s inductive bias and physical causality rather than mere architectural complexity. By establishing a standardized, leakage-free protocol and introducing metrics, this work provides a principled foundation for evaluating whether AI models can withstand the high-impact, tail-risk events that threaten modern grid stability.

Moving forward, R$^2$Energy serves as a catalyst for developing physically-augmented and climate-adaptive AI. It inspires a new generation of forecasting models that are not only computationally efficient but also capable of autonomously adapting their strategies when transitioning between stable regimes and high-entropy weather fronts. Ultimately, this benchmark bridges the gap between academic research and real-world deployment, guiding the development of trustworthy forecasting solutions essential for the global transition to a carbon-neutral energy future.